\documentclass{article}

     \usepackage[nonatbib,preprint]{neurips_2020}

\usepackage[utf8]{inputenc} 
\usepackage[T1]{fontenc}    
\usepackage{hyperref}       
\usepackage{url}            
\usepackage{booktabs}       
\usepackage{amsfonts}       
\usepackage{nicefrac}       
\usepackage{microtype}      
\usepackage{acronym}        
\usepackage{amssymb}
\usepackage{amsmath}
\usepackage{amsfonts}
\usepackage{caption}
\usepackage{subcaption}
\usepackage{mathtools}
\usepackage[ruled]{algorithm2e}
\title{Feature Whitening via Gradient Transformation\\ for Improved Convergence}
\author{%
  Shmulik Markovich-Golan\\
  \small{shmulik.markovich-golan@intel.com}
  \And Barak Battash\\
  \small{barak.battach@intel.com}
  \And Amit Bleiweiss\\
  \small{amit.bleiweiss@intel.com}
}

\def\Nl{{N_{l}}}
\def\Ml{{M_{l}}}

\def\vXpl{{\mathbf{X}_{l}^{p}}}

\def\vxn{{\mathbf{x}_{n}}}
\def\vxnp{{\mathbf{x}_{n}^{p}}}
\def\vXp{{\mathbf{X}^{p}}}

\def\vZp{{\mathbf{Z}^{p}}}

\def\vVxn{{\mathbf{V}_{x,n}}}

\def\vI{{\mathbf{I}}}
\def\diag{{\textrm{diag}}}

\def\vyn{{\mathbf{y}_{n}}}
\def\vynp{{\mathbf{y}_{n}^{p}}}
\def\vmux{{\mathbf{\mu}_{x}}}
\def\vPhix{{\mathbf{\Phi}_{x}}}
\def\vT{{\mathbf{T}}}
\def\vTT{{\mathbf{T}^{T}}}
\def\vVx{{\mathbf{V}_{x}}}
\def\vVxT{{\mathbf{V}_{x}^{T}}}
\def\vLambdax{{\mathbf{\Lambda}_{x}}}

\def\vzrp{{\mathbf{z}_{r}^{p}}}
\def\vWrn{{\mathbf{W}_{rn}}}
\def\vb{{\mathbf{b}}}
\def\vtWrn{{\tilde{\mathbf{W}}_{rn}}}
\def\vtb{{\tilde{\mathbf{b}}}}
\def\vPhiy{{\mathbf{\Phi}_{y}}}
\def\vPhiyT{{\mathbf{\Phi}_{y}^{T}}}
\def\vve{{\mathbf{v}_{e}}}
\def\vveT{{\mathbf{v}_{e}^{T}}}
\def\lambdae{{\lambda_{e}}}

\def\blambdax{{\bar{\lambda}_{x}}}

\def\tvT{{\tilde{\mathbf{T}}}}
\def\tvTT{{\tilde{\mathbf{T}}^{T}}}

\def\vxi{{\mathbf{\xi}}}
\def\vc{{\mathbf{c}}}

\def\cO{{\mathcal{O}}}
\def\cL{{\mathcal{L}}}

\def\vbWrn{{\bar{\mathbf{W}}_{rn}}}

\def\Ms{{M_s}}
\def\hMs{{\hat{M}_s}}
\def\gmax{{g_{\textrm{max}}}}
\def\vg{{\mathbf{g}}}

\def\ge{{g_{e}}}
\def\ae{{a_{e}}}
\def\phixm{{\phi_{x,m}}}
\def\vPhiyn{{\mathbf{\Phi}_{y,n}}}
\def\vtve{{\mathbf{t}_{ve}}}
\def\vtveT{{\mathbf{t}_{ve}^{T}}}
\def\vQ{{\mathbf{Q}}}
\def\vQs{{\mathbf{Q}_{s}}}
\def\vem{{\mathbf{e}_{m}}}
\def\vemT{{\mathbf{e}_{m}^{T}}}
\def\vemt{{\mathbf{e}_{m'}}}
\def\vemtT{{\mathbf{e}_{m'}^{T}}}

\acrodef{SGD}{stochastic gradient descent}
\acrodef{NGD}{natural gradient descent}
\acrodef{DNN}{deep neural network}
\acrodef{EVD}{Eigenvalue decomposition}
\acrodef{PCA}{principle component analysis}
\acrodef{ZCA}{zero-phase component analysis}
\acrodef{MAC}{multiply and accumulate}
\acrodef{BN}{batch norm}
\acrodef{AIC}{Akaike information criterion}
\acrodef{FIM}{Fisher information matrix}

\usepackage{graphicx}

\begin{document}

\maketitle

\vspace{-5 pt}
\begin{abstract}
Feature whitening~\cite{lecun2012efficient} is a known technique~ for speeding up training of \acs{DNN}. Under certain assumptions, whitening the activations~\cite{desjardins2015natural} reduces the \acl{FIM} to a simple identity matrix, in which case  \acl{SGD} is equivalent to the \emph{faster} \acl{NGD}. Due to the additional complexity resulting from transforming the layer inputs  and their corresponding gradients in the forward and backward propagation, and from repeatedly computing the \ac{EVD}, this method is not commonly used to date. 

 In this work, we address the complexity drawbacks of feature whitening. Our contribution is twofold. First, we derive an equivalent method,  which replaces the sample transformations by a transformation to the weight gradients, applied to every batch of $B$ samples. The complexity is reduced by a factor of $S/(2B)$, where $S$ denotes the feature dimension of the layer output. As the batch size increases with distributed training, the benefit of using the proposed method becomes more compelling. Second, motivated by the theoretical relation between the condition number of the sample covariance matrix and the convergence speed, we derive an alternative sub-optimal algorithm which recursively reduces the condition number of the latter matrix. Compared to \ac{EVD}, complexity is reduced by a factor of the input feature dimension $M$.   We exemplify the proposed algorithms with ResNet-based networks for image classification demonstrated on the CIFAR and Imagenet datasets. Parallelizing the proposed algorithms is straightforward and we implement a distributed version thereof. Improved convergence, in terms of speed and attained accuracy, can be observed in our experiments. 
\end{abstract}

\vspace{-5 pt}
\section{Introduction}
\label{sec:introduction}
\vspace{-5 pt}
Feature whitening is a known method for speeding-up training~\cite{le1991eigenvalues,wiesler2011convergence,raiko2012deep,lecun2012efficient}, which aims to decorrelate the inputs to the network layers, making it possible to match the learning rate at each eigenspace to its optimum value, inversely proportionate to the corresponding eigenvalue. Due to the additional complexity, resulting from transforming the layer inputs and their gradients in the forward- and backward-propagation stages, and from repeatedly computing the \ac{EVD}, using it is uncommon. The celebrated \ac{BN} method~\cite{ioffe2015batch} implements a degenerate version of feature whitening, aiming to standardize the features without decorrelating them. Although sub-optimal, \ac{BN} was found to speed up convergence~\cite{bjorck2018understanding} and has become a common practice in designing \acp{DNN} thanks to  its appealingly low complexity. In \cite{desjardins2015natural} it has been shown that when the features become white the \ac{FIM} reduces to an identity matrix and the simple \ac{SGD} optimization  coincides with the \ac{NGD} optimization~\cite{amari1998natural}  which is more suitable for difficult optimization tasks~\cite{povey2014parallel}. Some computations can be saved with the assumption that  feature statistics vary slowly during training, thus allowing to update the \ac{PCA}-based whitening transformation every block of samples and amortize the complexity of the \ac{EVD} computation. Incorporating \ac{BN} with feature whitening has been found to be useful to limit the variation of feature statistics within blocks. Instability of feature-whitening has been reported in~\cite{ioffe2015batch} and was attributed to treating the transformation as constant and independent of the data during the back propagation stage. In \cite{huang2018decorrelated} the stability issue was investigated, and it was found that constructing the whitening matrix based on \ac{ZCA}~\cite{bell1997independent, kessy2018optimal} is more stable than based on \ac{PCA} as it maintains a lower distortion compared to the original samples and avoids the stochastic axis swapping problem. Furthermore, the derivatives with respect to the whitening transformation were computed and incorporated in the back propagation stage, which further improved convergence speed, at the expense of increased complexity.

In this work, we consider the feature whitening method~\cite{desjardins2015natural} using \ac{ZCA} instead of \ac{PCA} and derive two reduced complexity methods. The first method is equivalent to the aforementioned method, and yet, computations are saved by transforming the weight gradients on every batch instead of whitening the activations and their gradients at every sample. The second method replaces the \ac{EVD} computation with an approximate recursive algorithm which conditions the activations' covariance matrix one subspace at a time. The structure of the paper is as follows: In Sec.~\ref{sec:direct_whitening} we present the feature whitening method of \cite{desjardins2015natural}. Then in Secs.~\ref{sec:gradient_whitening} and \ref{sec:recursive_whitening} we derive the proposed reduced complexity methods. The complexity of the proposed methods is analyzed in Sec.~\ref{sec:complexity}. We evaluate the proposed methods with various datasets and models in Sec.~\ref{sec:experimental_study} and conclude the work in Sec.~\ref{sec:conclusion}.

\vspace{-5 pt}
\section{Direct feature whitening}
\label{sec:direct_whitening}
\vspace{-5 pt}
We describe the feature whitening method as presented in \cite{desjardins2015natural}, where the transformation is constructed using \ac{ZCA} as in \cite{huang2018decorrelated}, and denote it by the \emph{direct} feature whitening method. 
%
Consider the input of the $l$-th layer of a \ac{DNN}, denoted by an $\Ml\times \Nl$ matrix $\vXpl$, where $p$ denotes the input index, $\Nl$ is the number of columns corresponding to the spatial dimension and $\Ml$ is the number of rows, corresponding to the feature dimension. This formulation is generic, where any spatial dimension can be vectorized (using the $\textit{vec}\left(\cdot\right)$ operator) into a single dimension.  \ac{SGD} is used to update the parameters, where it has been shown to be equivalent to the \ac{NGD} optimizer when the features are white and under certain assumptions~\cite{desjardins2015natural}. For brevity of notation, we omit the $l$-th layer index hereafter, unless explicitly stated.

Denote the $n$-th spatial element of the input, for $n=0,\ldots,N-1$, as $\vxnp$, and define the input matrix
\begin{align}
\textstyle \vXp \triangleq \left[\mathbf{x}_0^p, \mathbf{x}_1^p,\ldots,\mathbf{x}_{N-1}^p\right].\label{eq:vXp}
\end{align}

During the training procedure, as the network adapts, the mean and covariance of $\vxn$ vary as well and correspondingly the whitening matrix has to be adapted too.
A common method for continuously tracking the moments is to use the empirical sample-mean and sample-covariance-matrix, computed  in blocks of $L$ batches, each consisting of $B$ samples, and recursively average them between blocks. The sample indices of the $i$-th block are $p\in\left[LBi, LB(i+1)-1\right]$. The varying mean and covariance are estimated at block index $i\geq1$ by:
\begin{align}
\vmux\left(i\right) \triangleq& \alpha\vmux\left(i-1\right)+\left(1-\alpha\right)\frac{1}{LBN}\sum_{p=LBi}^{LB(i+1)-1}\sum_{n}\vxnp\label{eq:vmuxi}\\
\vPhix\left(i\right) \triangleq& \alpha\vPhix\left(i-1\right)+\left(1-\alpha\right)\frac{1}{LBN}\sum_{p=LB i}^{LB(i+1)-1}\sum_{n}\left(\vxnp-\vmux\left(i\right)\right)\left(\vxnp-\vmux\left(i\right)\right)^{T}\label{eq:vPhixi}
\end{align}
where $0\leq\alpha<1$ is a recursive-averaging factor between blocks.
The mean and  covariance are initialized in the first block ($i=0$) to:
\begin{align}
\vmux\left(0\right) \triangleq& \frac{1}{LBN}\sum_{p=0}^{LB-1}\sum_{n}\vxnp\label{eq:vmuxnz}\\
\vPhix\left(0\right) \triangleq& \frac{1}{LBN}\sum_{p=0}^{LB-1}\sum_{n}\left(\vxnp-\vmux\left(0\right)\right)\left(\vxnp-\vmux\left(0\right)\right)^{T}.\label{eq:vPhixnz}
\end{align}
Assuming that the first and second order moments vary slowly and that they are similar across space, we can reduce the estimation complexity by averaging over a subset of the samples and of spatial elements.
Storing the moments requires additional $M(M+1)$ memory elements.

We turn to the construction of the whitening matrix at the $i$-th block. A whitening matrix $\vT(i)$  satisfies the following condition:
\begin{align}
\vT(i)\vPhix(i)\vTT(i)=\vI\label{eq:whitening}
\end{align}
where $\vI$ is an $M\times M$ identity matrix. The whitening matrix $\vT(i)$ is not unique and can be constructed in several methods~\cite{golub2012matrix}. Some methods are based on the \ac{EVD} of $\vPhix(i)$, defined as:
\begin{align}
\textstyle \vPhix(i) = \vVx(i)\vLambdax(i)\vVxT(i)\label{eq:vPhix_evd}
\end{align}
where $\vVxn(i)$ is a $M\times M$ dimensional eigenvectors matrix,
\begin{align}
\textstyle \vLambdax(i) \triangleq \diag\left(\left[\lambda_{x,0}(i),\cdots,\lambda_{x,M-1}(i)\right]\right)\label{eq:vLambdax}
\end{align}
is a diagnonal $M\times M$ dimensional eigenvalues matrix and $\diag\left(\cdot\right)$ denotes a diagonal matrix with the argument vector placed at the diagonal. Without loss of generality we assume that the eigenvalues are sorted in decreasing order, such that $\lambda_{x,0}(i)\geq\lambda_{x,1}(i)\geq\cdots\geq\lambda_{x,M-1}(i)$. Given the latter decomposition, the \ac{ZCA}-based whitening matrix can be constructed by:
\begin{align}
\textstyle \vT(i) \triangleq \vVx(i)\diag^{1/2}\left(\vg(i)\right)\vVxT(i)\label{eq:vT_EVD}
\end{align}
where $\diag^{1/2}\left(\vg\left(i\right)\right)\triangleq\diag\left(\left[\sqrt{g_0(i)},\ldots,\sqrt{g_{M-1}(i)}\right]^{T}\right)$,  the gains vector $\vg$ is defined as
\begin{align}
\vg(i) \triangleq \left[{1}/{\max\left\{\lambda_{x,0}(i),\epsilon\right\}},\cdots,{1}/{\max\left\{\lambda_{x,M-1}(i),\epsilon\right\}}\right]^{T}\label{eq:vg}
\end{align}
and $\epsilon>0$ limits the minimal denominator for numerical stability. 

Denote the output of the direct feature whitening transformation in the forward-propagation stage of the $i$-th block as:
\begin{align}
\textstyle \vynp \triangleq \vT(i-1)\left(\vxnp-\vmux(i-1)\right)\label{eq:vyn}
\end{align}
Note that $i$-th block outputs are constructed using the $i-1$-th transformation. The whitening transformation is initialized as:
\begin{align}
\textstyle \vT(-1) \triangleq& \vI\label{eq:vT_init}\\
\textstyle \vmux(-1) \triangleq& \mathbf{0}\label{eq:vmux_init}
\end{align}
In the back propagation stage, propagating the gradient through the whitening transformation from \eqref{eq:vyn} requires computing:
\begin{align}
\frac{\partial\cL}{\partial\vxnp} = \frac{\partial\vynp}{\partial\vxnp} \frac{\partial\cL}{\partial\vynp}  = \vTT(i-1)\frac{\partial\cL}{\partial\vynp}\label{eq:dxn}
\end{align}
where $\cL$ denotes the loss-function that is being optimized. Note that we adopt the denominator layout notation convention when considering derivatives with respect to vectors or matrices.

Denote the $S\times R$ dimensional output of the $l$-th layer as $\vZp$, where $S$ and $R$ respectively denote the feature and space dimensions. As in \cite{desjardins2015natural},  we limit the discussion to whitening the input of linear layers. In this case the $l$-th layer can be formulated as
\begin{align}
\vzrp \triangleq \sum_{n}\vWrn\vynp+\vb\label{eq:vzrp}
\end{align}
for $r=0,\ldots,R-1$, where the trained parameters of the layer are $S\times M$ dimensional weight matrices $\left\{\vWrn\right\}_{rn}$, for $r\in\left[0,R-1\right]$ and $n\in\left[0,N-1\right]$, and a $S\times 1$ dimensional bias vector $\vb$.
Convolution and fully-connected layers can be formulated as special cases of this generic formulation. Substituting the transformed input \eqref{eq:vyn} into \eqref{eq:vzrp} yields:
\begin{align}
\vzrp = \sum_{n}\vWrn\vT(i-1)\left(\vxnp-\vmux(i-1)\right)+\vb.\label{eq:vzrp2}
\end{align}

Considering \eqref{eq:vzrp},  the gradient with respect to the weights and bias in the $j$-th batch is computed  by:
\begin{align}
\frac{\partial\cL}{\partial \vWrn}=&\sum_{p=jB}^{(j+1)B-1}\sum_{s}\frac{\partial z_{rs}^{p}}{\partial\vWrn}\frac{\partial\cL}{\partial z_{rs}^{p}}=\sum_{p=jB}^{(j+1)B-1}\frac{\partial \cL}{\vzrp}\left(\vynp\right)^{T}\label{eq:dWrn}\\
\frac{\partial\cL}{\partial \vb}=&\sum_{p=jB}^{(j+1)B-1}\sum_{r,s}\frac{\partial z_{rs}^{p}}{\partial\vb}\frac{\partial\cL}{\partial z_{rs}^{p}}=\sum_{p=jB}^{(j+1)B-1}\sum_{r}\frac{\partial \cL}{\vzrp}\label{eq:db}
\end{align}
for $r=0,\ldots,R-1$ and $n=0,\ldots,N-1$. 
The parameters are updated on every batch according to the \ac{SGD} optimization rule:
\begin{align}
\vWrn\coloneqq&\vWrn-\eta\frac{\partial\cL}{\partial\vWrn}\label{eq:vWrn_SGD}\\
\vb\coloneqq&\vb-\eta\frac{\partial\cL}{\partial\vb}\label{eq:vb_SGD}
\end{align}
where $\coloneqq$ denotes the assignment operator and $\eta$ denotes the learning rate.

\vspace{-5 pt}
\section{Gradient-based feature whitening}
\label{sec:gradient_whitening}
\vspace{-5 pt}
We propose the gradient-based feature whitening method in Sec.~\ref{subsec:gradient_whitening_method} and prove its equivalence to the direct feature whitening method from the previous section. In Sec.~\ref{subsec:gradient_whitening_practice} we present some practical considerations and supplements to the basic method.
\vspace{-5 pt}
\subsection{Method}
\label{subsec:gradient_whitening_method}
\vspace{-5 pt}
The  forward propagation stage of the method is given by:
\begin{align}
\vzrp = \sum_{n}\vtWrn\left(\vxnp-\vmux\right)+\vtb\label{eq:vzrp3}
\end{align}
where $\left\{\vtWrn\right\}_{r,n}$ for $r\in\left[0,R-1\right]$ and $n\in\left[0,N-1\right]$ and $\vtb$ are the trained parameters.
Unlike in \cite{desjardins2015natural}, $\vtWrn$ is not split into two parts, and applied as a single transformation, conceptually including both the whitening and trained transformations. The activations and their gradients in the forward  and backward propagation are computed normally, and do not require another whitening transformation as in~\cite{desjardins2015natural}.
The whitening matrix is computed as in \eqref{eq:vT_EVD}, however, it is only used in the back propagation stage to construct the weight gradient transformation, such that the weights are updated as in the direct whitening method.

The equivalence is proven by induction. Without loss of generality, let us assume that \eqref{eq:vzrp3} is equivalent to the forward propagation of the direct whitening in \eqref{eq:vzrp2}, and that the corresponding whitening transformations are identical, i.e.:
\begin{align}
\sum_{n}\vtWrn\left(\vxnp-\vmux\right)+\vtb = \sum_{n}\vWrn\vT\left(\vxnp-\vmux\right)+\vb
\end{align}
for $r=0,\ldots,R-1$. It follows that:
\begin{align}
\vtWrn=&\vWrn\vT\label{eq:vtWrn_rel}\\
\vtb=&\vb.\label{eq:vtb_rel}
\end{align}
We consider the relations between the parameter gradients in the direct whitening method, denoted $\frac{\partial\cL}{\partial\vWrn}$ and $\frac{\partial\cL}{\partial\vb}$, and their gradient-based whitening method counterparts, denoted $\frac{\partial\cL}{\partial\vtWrn}$ and $\frac{\partial\cL}{\partial\vtb}$. Considering \eqref{eq:vtWrn_rel}, \eqref{eq:vtb_rel} and using matrix derivative rules, we obtain the relations:
\begin{align}
\frac{\partial\cL}{\partial\vWrn}=&\frac{\partial\cL}{\partial\vtWrn}\vTT\label{eq:dvtWrn_rel}\\
\frac{\partial\cL}{\partial\vb}=&\frac{\partial\cL}{\partial\vtb}.\label{eq:dvb_rel}
\end{align}
Finally, following the \ac{SGD} update rules in \eqref{eq:vWrn_SGD} and \eqref{eq:vb_SGD} and substituting  \eqref{eq:vtWrn_rel} and \eqref{eq:vtb_rel}, we derive the update rules of the gradient-based whitening as:
\begin{align}
\vtWrn\coloneqq&\vtWrn-\eta\frac{\partial\cL}{\partial\vbWrn}\label{eq:vtWrn_SGD}\\
\vtb\coloneqq&\vtb-\eta\frac{\partial\cL}{\partial\vtb}\label{eq:vtb_SGD}\\
\frac{\partial\cL}{\partial\vbWrn} \triangleq&  \frac{\partial\cL}{\partial\vtWrn}\vQ\label{eq:mod_weight_grad}\\
\vQ\triangleq&\vTT\vT.\label{eq:vQ}
\end{align}
Note that these update rules satisfy that the updated parameters in the gradient-based whitening method are equal to the direct whitening method counterparts, and therefore the varying networks are identical and the methods are equivalent  by construction.

Special treatment is required for every $i$-th block, when the empirical mean is updated from $\vmux(i-1)$ to $\vmux(i)$. In these cases, the bias parameter $\vtb$ is updated to compensate for the update and maintain the overall network unchanged after update.  Considering \eqref{eq:vzrp3} and respectively denoting by $\vtb|_{i^-}$ and $\vtb|_{i}$ the bias prior to and after the update, we would like to maintain the relation ${1}/{R}\sum_{r,n}\vtWrn\left(\vxnp-\vmux(i)\right)+\vtb|_{i}={1}/{R}\sum_{r,n}\vtWrn\left(\vxnp-\vmux(i-1)\right)+\vtb|_{i^-}$. The bias compensation rule is therefore defined as:
\begin{align}
\vtb|_{i} \triangleq \vtb|_{i^-}+{1}/{R}\sum_{r,n}\vtWrn\left(\vmux\left(i-1\right)-\vmux\left(i\right)\right).\label{eq:vtb_corr}
\end{align}
This concept is similar to  \cite{desjardins2015natural}. However, there is no need to compensate for changes in the whitening matrix $\vT$, as they are only manifested in the gradients and not explicitly applied to the activations.
\vspace{-5 pt}
\subsection{Practical considerations}
\label{subsec:gradient_whitening_practice}
\vspace{-5 pt}
We present some modifications to the basic method from the previous section, which we found to improve performance and robustness in practice. In the general case the $M$ dimensional feature subspace can be split into a signal subspace which corresponds to the $\Ms$ largest eigenvalues and holds most of the energy and a noise subspace which corresponds to the smallest $M-\Ms$ eigenvalues. We conjecture that the noise subspace does not contain information that is instrumental to further reduction of the loss function. Consequently, amplifying the noise components might have a destructive effect and hamper the convergence process. We propose to adopt the \ac{AIC}~\cite{akaike1974new} for estimating the rank of the signal subspace $\Ms$:
\begin{align}
\hMs \triangleq& \sum_m\exp\left(-\bar{\lambda}_m\log\bar{\lambda}_m\right)\label{eq:hMs}\\
\bar{\lambda}_m \triangleq& {\lambda_m}/{\sum_{m'}{\lambda_{m'}}}
\end{align}
where the latter are the normalized eigenvalues for $m=0,\ldots,M-1$. Note that in the extreme case of identical eigenvalues,  we obtain $\hMs=M$. In the other extreme case where the samples are completely correlated, i.e., $\lambda_0>0$ and $\lambda_m\rightarrow0$ for $m=1,\ldots,M-1$, we obtain $\hMs\rightarrow1$. 

Instead of forcing the variance of the whitened signals to be $1$, we maintain the average variance at the input, and limit the maximal gain to $\gmax$. To summarize, the gain in \eqref{eq:vg} is replaced by:
\begin{align}
g_m = \min\left\{1/(M\bar{\lambda}_m), \gmax\right\} \label{eq:g_mod}
\end{align}
for $m=0,\ldots,\hMs-1$ and $1$ otherwise.
Finally, we suggest to use a recursively smoothed version of $\vQ(i)$, denoted as $\vQs(i)$, initialized by $\vQs(-1)\triangleq\vI$ and updated every block according to:
\begin{align}
\vQs(i) = \beta\vQs(i-1)+(1-\beta)\vQ(i)\label{eq:vQs}
\end{align}
where $0\leq\beta<1$ is a recursive averaging factor.


\vspace{-5 pt}
\section{Recursive feature whitening}
\label{sec:recursive_whitening}
\vspace{-5 pt}
Haykin~\cite{haykin2005adaptive} analyzed the stability and convergence of the gradient descent algorithm for Gaussian signals and a linear system. Similarly to LeCun~\cite{lecun2012efficient} he concluded that the convergence time is proportionate to the condition number of the input covariance matrix, i.e., convergence-time is shorter when the eigenvalues spread is reduced. At the limit, when the condition number equals $1$, the convergence time is minimal and the corresponding input covariance matrix is white.

Motivated by the high computational complexity of \ac{EVD}, and by the relation between the convergence time and the condition number of the input covariance matrix, we propose a recursive approach for reducing the condition number. The idea of the recursive approach is to monitor the covariance matrix of the transformed input $\vPhiyn(i)$, computed similarly to \eqref{eq:vPhixi}, and in every block identify a \emph{high power} subspace. Then, construct a simple transformation step to reduce its power and append it to the previous whitening transformation, thereby reducing the condition number of $\vPhiyn(i+1)$ in the following block.  The method for constructing the transformation is described in Sec.~\ref{subsec:rec_whitening_method}. For estimating the high power subspace, we refer to \cite{barnov2017qrd} and adopt the procedure for estimating the principal eigenvector, which we present in  Sec.~\ref{subsec:principal_ev}. 
\vspace{-5 pt}
\subsection{Method}
\label{subsec:rec_whitening_method}
\vspace{-5 pt}
Let $\vve(i)$ denote a high power subspace of $\vPhiy\left(i\right)$, containing more power than the average eigenvalue. The vector $\vve(i)$ is normalized, such that $\|\vve\|=1$.
The power contained in its subspace is denoted $\lambdae(i)$ and is computed by:
\begin{align}
\lambdae(i) \triangleq \vveT(i)\vPhiy(i)\vve(i).\label{eq:lambda}
\end{align}
Define the recursive average power of the $m$-th input feature as:
\begin{align}
\phixm\left(i\right) \triangleq& \alpha\phixm\left(i-1\right)+\left(1-\alpha\right)\frac{1}{LBN}\sum_{p=LBi}^{LB(i+1)-1}\sum_{n}\left(x_{n,m}^{p}\right)^2\label{eq:phixmi}
\end{align}
and define the average over all input features as:
\begin{align}
\blambdax(i) = 1/M\sum_m\phixm(i).\label{eq:blambday}
\end{align}
We define the transformation-step which reduces the power in the subspace of $\vve$ as a summation of two projection matrices:
\begin{align}
\tvT(i) \triangleq& \sqrt{\ge(i)}\vve(i)\vveT(i)+\left(\vI-\vve(i)\vveT(i)\right)\nonumber\\
=& \vI+(\sqrt{\ge(i)}-1)\vve(i)\vveT(i)=\vI+\ae(i)\vve(i)\vveT(i)\label{eq:tvT}\\
\ge(i) \triangleq& \frac{\delta\blambdax(i)}{\max\left\{\lambdae(i), \epsilon\right\}}\label{eq:ge}\\
\ae(i)\triangleq&\sqrt{\ge(i)}-1\label{eq:xi}
\end{align}
with $\delta>0$ being a parameter controlling the power reduction. Note that the transformation step reduces the power of the high power subspace to $\vveT(i)\tvTT(i)\vPhiy(i)\tvT(i)\vve(i)=\delta\blambdax(i)$, whereas all orthogonal subspaces remain unchanged.

Due to variations of the feature covariance matrix, resulting from variations of the network parameters, it is necessary to introduce a \emph{forgetting} procedure which reverts the transformation in subspaces which no longer contain high power. The procedure is based on \emph{leaking} a fraction of the input $\vxn$ to the whitened input $\vyn$. Given the previous block whitening transformation $\vT(i-1)$ and the recursive transformation-step \eqref{eq:tvT}, the update rule for the whitening transformation is defined as:
\begin{align}
\vT(i)\triangleq\gamma\tvT(i)\vT(i-1)+\left(1-\gamma\right)\vI=\gamma\left(\ae(i)\vve(i)\vtveT(i)+\vT(i-1)\right)+(1-\gamma)\vI\label{eq:vT_rec}
\end{align}
where $0<\gamma\leq 1$ is the leakage factor and
\begin{align}
\vtve(i)\triangleq&\vT(i-1)\vve(i).\label{eq:vtve}
\end{align}

Considering the transformation step~\eqref{eq:tvT} and \eqref{eq:vT_rec}, the matrix $\vQ(i)$ from \eqref{eq:vQ} can be efficiently computed as:
\begin{align}
\vQ(i) =& \gamma^2\left(\vQ(i-1)+\ae(i)\left(\ae(i)+2\right)\vtve(i)\vtveT(i)\right)+\left(1-\gamma\right)^2\vI\nonumber\\
&+\gamma(1-\gamma)\left(\ae(i)\left(\vve(i)\vtveT(i)+\vtve(i)\vveT(i)\right)+\vT(i-1)+\vTT(i-1)\right)\label{eq:vQ_rec}
\end{align}
requiring $\cO(M^2)$ \ac{MAC} operations instead of $\cO(M^3)$.
Similarly to \eqref{eq:vQs} in the direct feature whitening method, the weight gradients are transformed by an inter-block smoothed version  of \eqref{eq:vQ_rec}.

\vspace{-5 pt}
\subsection{High power subspace estimation}
\label{subsec:principal_ev}
\vspace{-5 pt}
We adopt the procedure from \cite{barnov2017qrd} for estimating the principal eigenvector, in the high signal-to-noise-ratio case. The block index $i$  is omitted for brevity.

Let $\vxi$ be an $M\times1$ vector comprised of the norms of the columns of $\vPhiy$, i.e.:
\begin{align}
\xi_{m} \triangleq \|\vPhiy \mathbf{e}_{m}\|\label{eq:xim}
\end{align}
for $m=0,1,\ldots,M-1$ where $\mathbf{e}_m\triangleq\left[\mathbf{0}_{1\times m-1}, 1, \mathbf{0}_{1\times M-m}\right]^{T}$ is a selection vector, used to pick the $m$-th column of $\vPhiy$.
Let
\begin{align}
m' \triangleq \textrm{argmax}_{m}\left(\left\{\xi_m\right\}_{m=0}^{M-1}\right)\label{eq:mt}
\end{align}
be the index of the column with the highest norm. The following method \emph{aligns} the other columns with the $m'$-th column, and the high power subspace is computes as the normalized average of the aligned columns.
Define the inner product between the $m'$-th column and every column in $\vPhiy$ as
\begin{align}
\vc^{T} \triangleq \mathbf{e}_{m'}^{T}\vPhiyT\vPhiy.\label{eq:vc}
\end{align}
Next, we define a set of column indices $\mathcal{M}$. An index $m$ is included in the set $\mathcal{M}$ if the correlation between its corresponding column and the $m'$-th column is sufficiently large, i.e.:
\begin{align}
|c_m|\geq\max\left\{c_{\textrm{rel}}\|\vPhiy\vemt\|\cdot\|\vPhiy\vem\|),c_{\textrm{abs}}\right\}
\end{align}
where $c_{\textrm{rel}}$ and $c_{\textrm{abs}}$ are \emph{relative} and \emph{absolute} coherence threshold.
Finally, we align the columns of $\vPhiy$ which indices are in $\mathcal{M}$, average them and normalize to obtain the high power subspace $\vve$:
\begin{align}
\tilde{\mathbf{v}}_{e} \triangleq& \frac{1}{|\mathcal{M}|}\sum_{m\in\mathcal{M}}\frac{1}{c_{m}}\vPhiy\mathbf{e}_{m}\label{eq:tve}\\
\vve \triangleq& \frac{1}{\|\tilde{\mathbf{v}}_{e}\|}\tilde{\mathbf{v}}_{e}.
\end{align}
%

\vspace{-5 pt}
\section{Complexity analysis}
\label{sec:complexity}
\vspace{-5 pt}
The  computational complexity of the direct feature whitening from Sec.~\ref{sec:direct_whitening}  is comprised of: 1) Computing the transformation at every block based on \ac{EVD}, requiring $M^3/(LB)$ \acp{MAC} per sample; 2) Transforming the input in the forward propagation stage, requiring $NM^2$ \acp{MAC} per sample; and 3) Transforming the activation gradients in the backward propagation stage, requiring $NM^2$ \acp{MAC} per sample. A total of $M^3/(LB)+2NM^2$ \acp{MAC} per sample are required.

The gradient-based feature whitening in Sec.~\ref{sec:gradient_whitening} replaces transforming of the activations and their gradients at each sample by instead transforming the weights gradients at each batch, requiring $N S M^2/B$ \acs{MAC} per sample. The computational complexity in the forward and backward propagation is therefore reduced by a factor of $S/(2B)$, which becomes lower with the tendency of the batch size to increase when training is performed in parallel over multiple machines. In case that $S/(2B)>1$, one could apply the direct whitening method.

In the recursive gradient-based whitening method in Sec.~\ref{sec:recursive_whitening}, computing the \ac{EVD} is replaced by recursively computing a \emph{high power} subspace and updating the transformation.  Also, by leveraging the structure of the recursive transformation, the gradient transformation $\vQ(i)$ is efficiently computed by $\cO(M^2)$ \acp{MAC} according to \eqref{eq:vQ_rec} instead of $\cO(M^3)$ in the generic matrix multiplication case. The computational complexity of constructing the whitening transformation is reduced by a factor $M$.

Note that we neglect the complexity of tracking the moments, which is identical in all methods. This is a reasonable assumption given there is a low variability of the statistics over consecutive activations  and different space elements.

\vspace{-5 pt}
\section{Experimental study}
\label{sec:experimental_study}
\begin{figure}[htb]
\vspace{-10 pt}
 \centering
 \begin{subfigure}[b]{0.32\textwidth}
 \centering
\includegraphics[width=\textwidth]{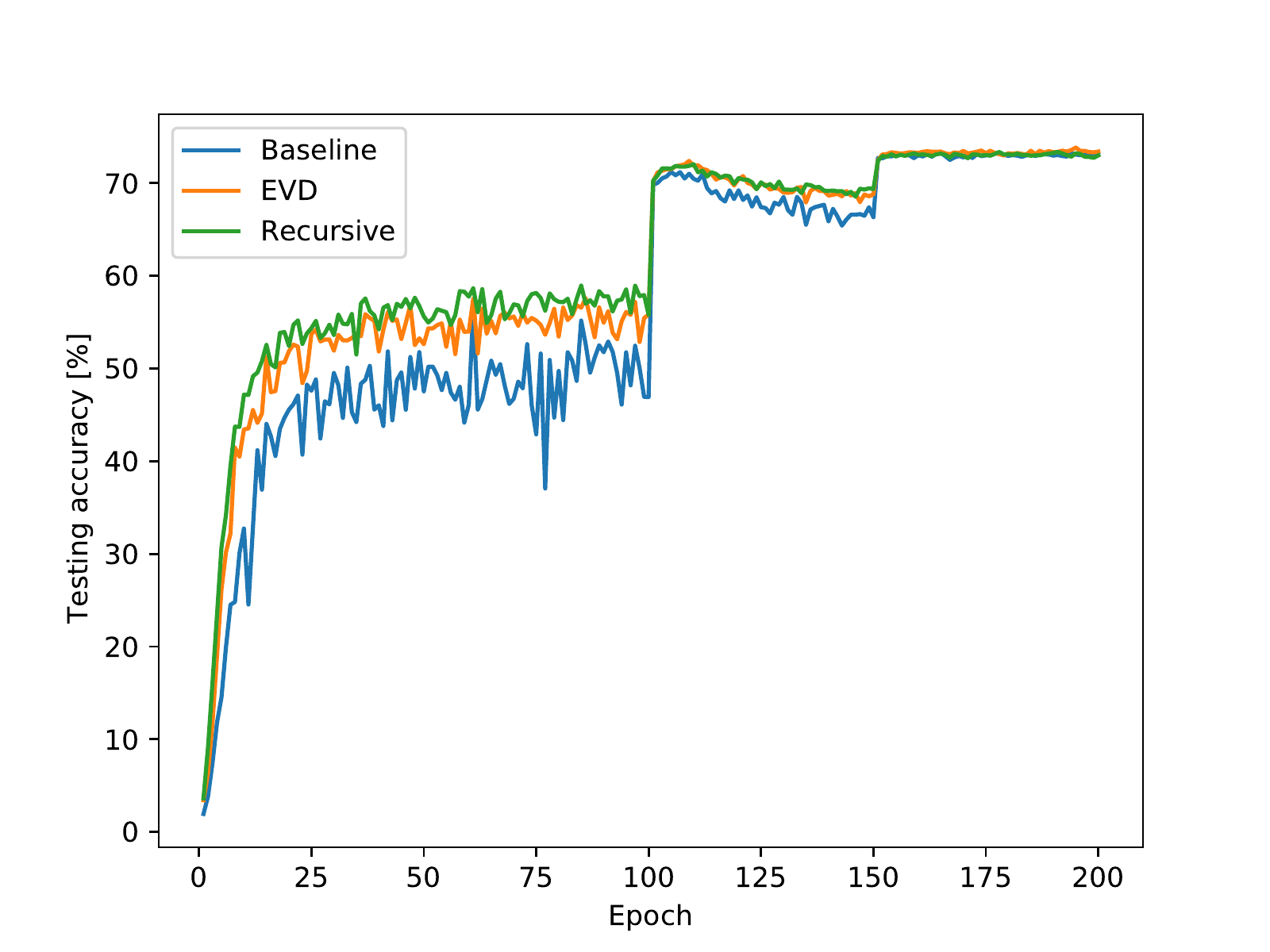}
 \caption{Testing accuracy}
 \label{fig:resnet110_accuracy}
 \end{subfigure}
 \hfill
 \begin{subfigure}[b]{0.32\textwidth}
 \centering
\includegraphics[width=\textwidth]{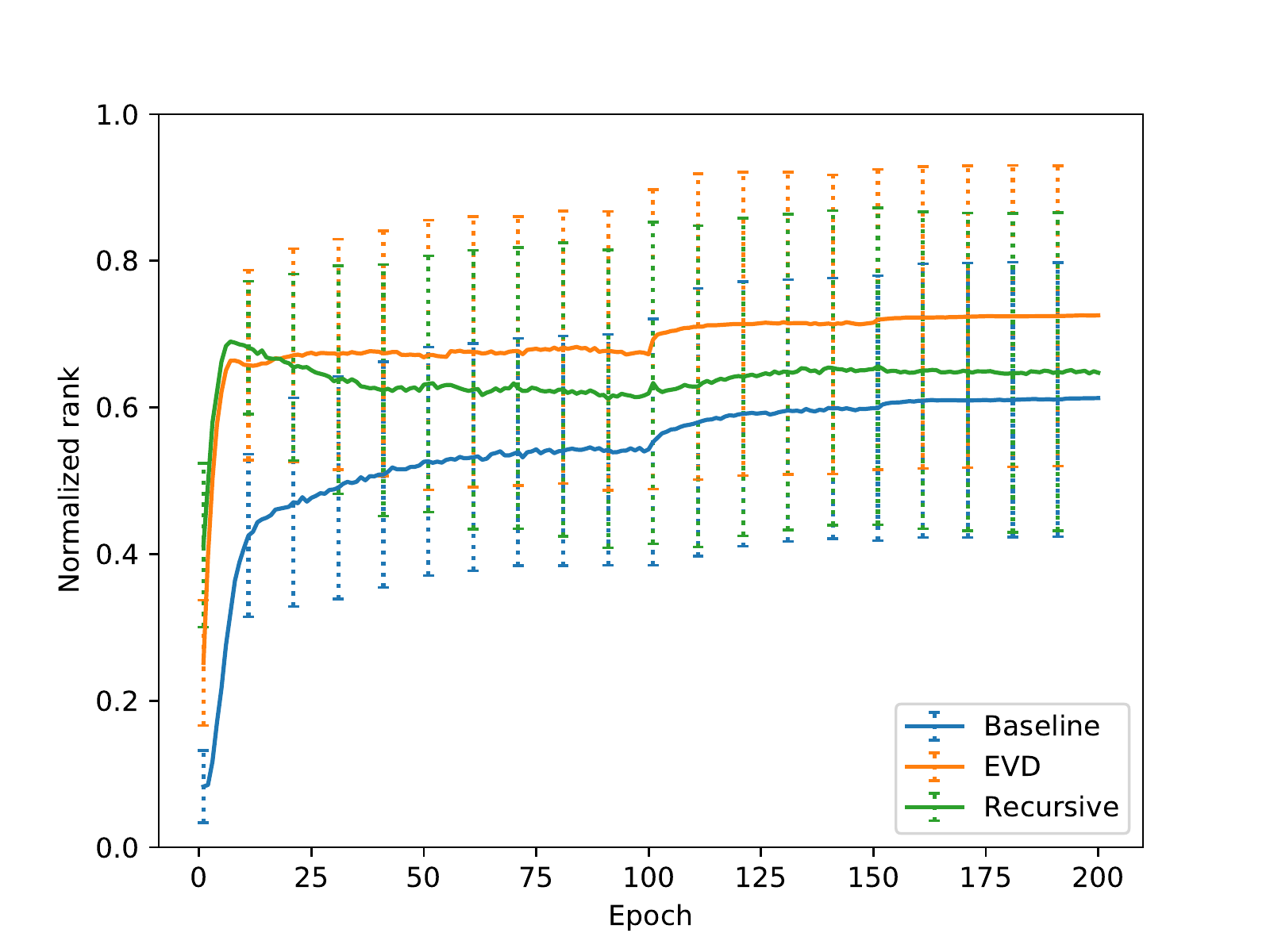}
 \caption{Normalized rank}
 \label{fig:resnet110_kappa}
 \end{subfigure}
 \hfill
  \begin{subfigure}[b]{0.32\textwidth}
 \centering
\includegraphics[width=\textwidth]{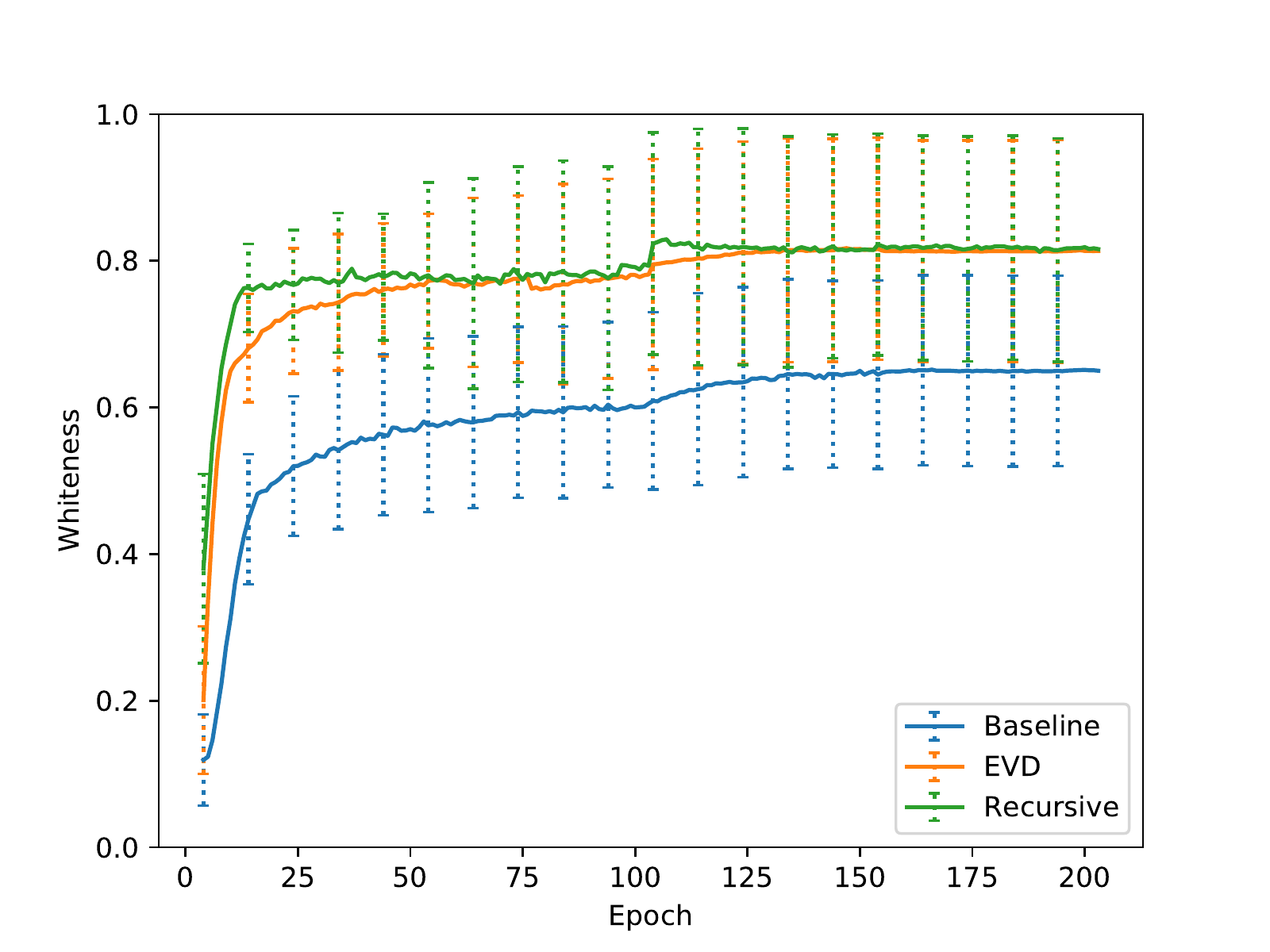}
\caption{Whiteness}
 \label{fig:resnet110_rho}
 \end{subfigure}
 \vspace{-5 pt}
\caption{Performance criteria vs. epoch number for ResNet-110 models over CIFAR}
\end{figure}

\begin{figure}[htb]
\vspace{-10 pt}
 \centering
 \begin{subfigure}[b]{0.32\textwidth}
 \centering
\includegraphics[width=\textwidth]{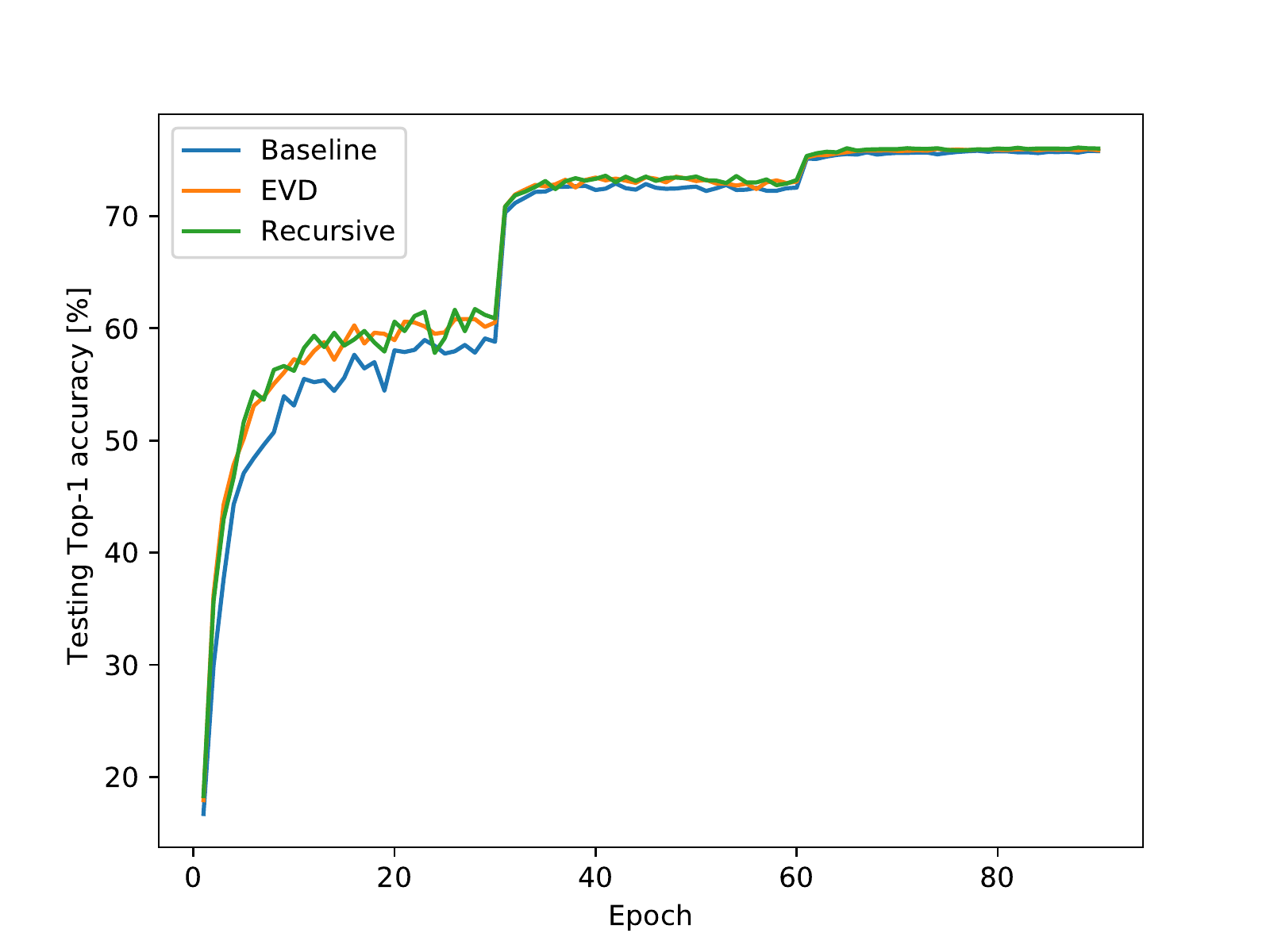}
 \caption{Testing accuracy}
 \label{fig:resnet50_accuracy}
 \end{subfigure}
 \hfill
 \begin{subfigure}[b]{0.32\textwidth}
 \centering
\includegraphics[width=\textwidth]{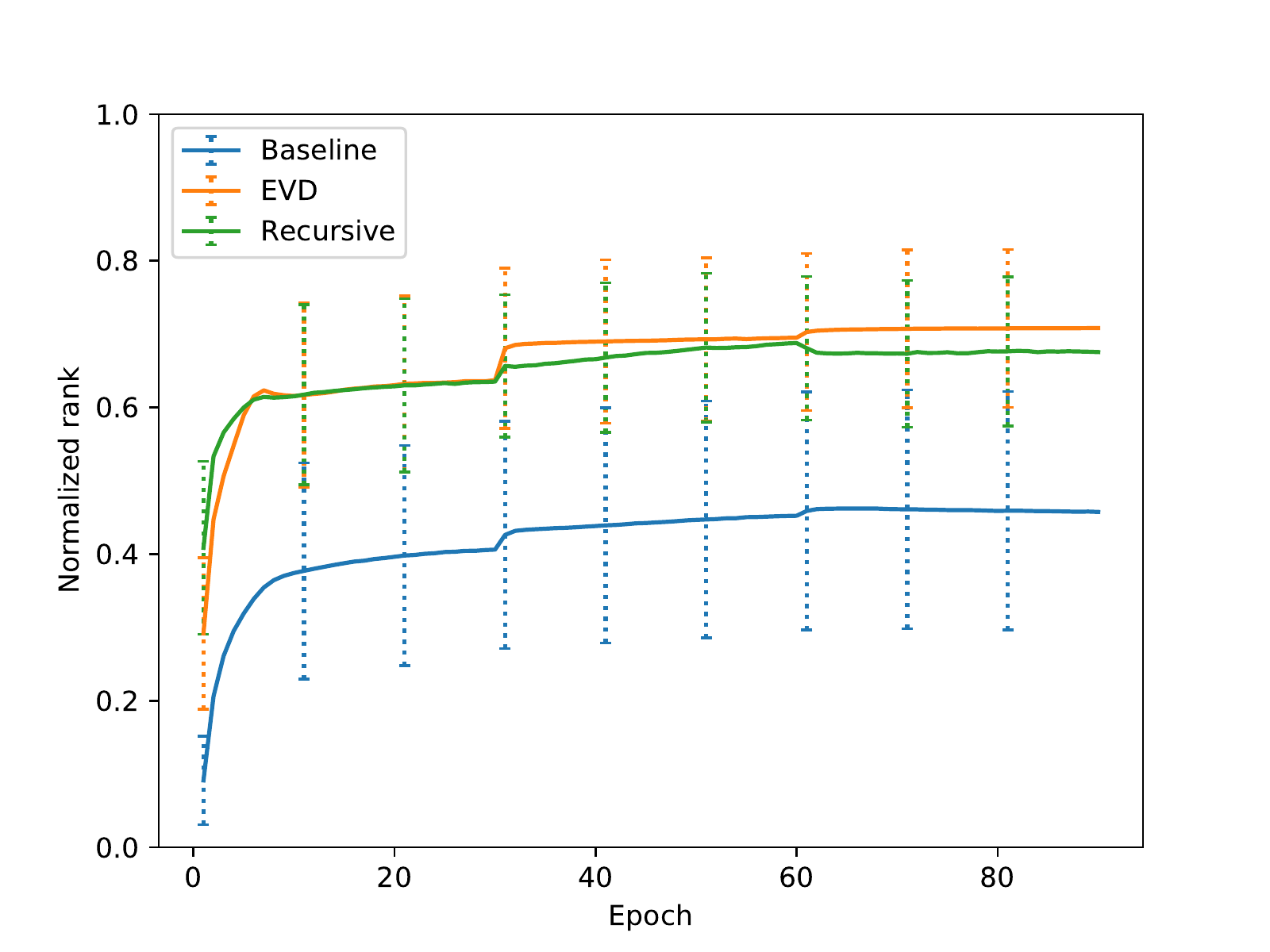}
 \caption{Normalized rank}
 \label{fig:resnet50_kappa}
 \end{subfigure}
 \hfill
 \begin{subfigure}[b]{0.32\textwidth}
 \centering
\includegraphics[width=\textwidth]{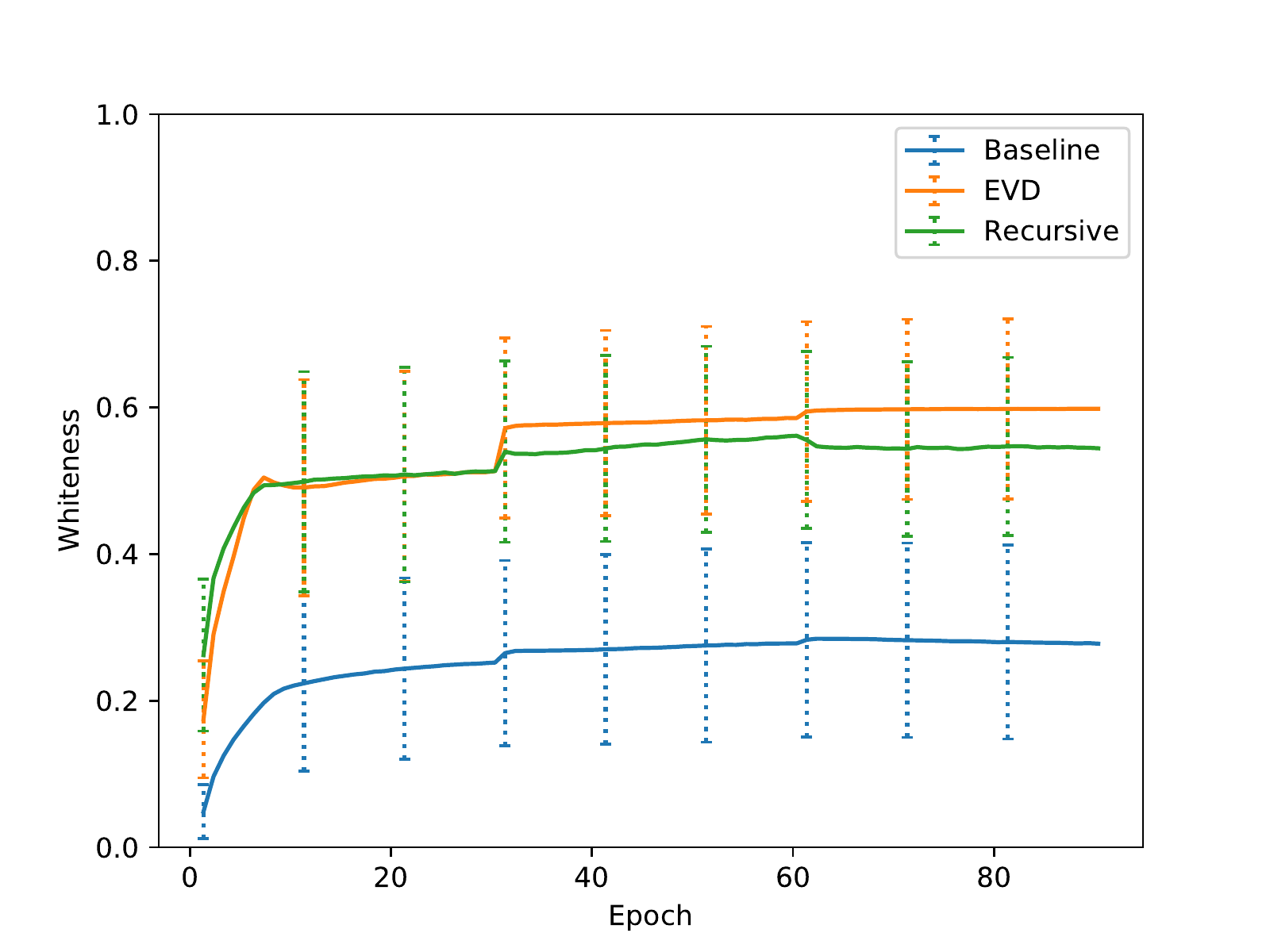}
 \caption{Whiteness}
 \label{fig:resnet50_rho}
 \end{subfigure}
 \vspace{-5 pt}
\caption{Performance criteria vs. epoch number for ResNet-50 models over Imagenet}
\end{figure}

\vspace{-5 pt}
We incorporate the proposed methods into the ResNet-110 and ResNet-50 models~\cite{he2016deep} and evaluate their convergence when training on the CIFAR~\cite{krizhevsky2009learning} and Imagenet~\cite{imagenet_cvpr09} datasets with $100$ and $1000$ classes, respectively. For each model we compare three methods: the original model, denoted as \emph{baseline}; the \ac{EVD} gradient-based feature whitening method, denoted as \emph{EVD}; and the recursive feature whitening method, denoted as \emph{recursive}. In models incorporating feature whitening we place a whitening layer prior to the convolution layers at each basic block (i.e., in ResNet-110 and ResNet-50 we respectively add $108$ and $52$ whitening layers). Training on CIFAR consists of $200$ epochs, where the learning rate is initialized to $0.1$, and reduces to $0.01$ and $0.001$ at epochs $100$ and $150$, respectively. The \ac{SGD} rule is applied with a momentum of $0.9$ and weight decay of $5e-4$. Training on Imagenet consists of $90$ epochs, where the learning rate is initialized to $0.1$, and reduces to $0.01$ and $0.001$ at epochs $30$ and $60$, respectively. The \ac{SGD} rule is applied with a momentum of $0.9$ and weight decay of $1e-4$. The hyper-parameters of the \ac{EVD}-based feature whitening are $\alpha=0.9$, $\beta=0.95$, $\gmax=10$ and $\epsilon=1e-5$. The hyperparameters of the recursive feature whitening are $\alpha=0.1$, $\beta=0.1$, $\gamma=0.99$, $\delta=0.25$,  $c_\textrm{rel}=0.025$, $c_\textrm{rel}=1e-6$ and $\epsilon=1e-5$.  In order to evaluate the effectiveness of feature whitening we evaluate the classification accuracy and examine the normalized rank and \emph{whiteness} of the covariance matrix of the whitened features. The normalized rank is defined as $\kappa \triangleq \hMs/M$, where $\hMs$ is estimated using \ac{AIC}. The \emph{whiteness} of the covariance matrix measures how \emph{close} it is to being diagonal, and is defined as
\begin{align}
    \rho \triangleq M/\sum_{m\neq m'} \frac{\left(\vemtT \vPhiy\vem\right)^2}{\vemT \vPhiy\vem\cdot\vemtT \vPhiy\vemt}\label{eq:rho}
\end{align}
where in the extreme case of a diagonal $\vPhiy$ we get $\rho=1$, and in the other extreme case of $\vyn$ being completely correlated we get $\rho=\frac{1}{M}$. We average the $\kappa$ and $\rho$ over all layers.
The testing accuracy, the average normalized rank and average whiteness of features covariance for ResNet-110 and ResNet-50 are respectively depicted in Figs.~\ref{fig:resnet110_accuracy}, \ref{fig:resnet110_kappa}, \ref{fig:resnet110_rho} and Figs.~\ref{fig:resnet50_accuracy}, \ref{fig:resnet50_kappa}, \ref{fig:resnet50_rho}. We  capture the variability of the normalized rank and whiteness across different whitening layers, and also depict its standard-error boundaries.
As expected, from the normalized rank Figs.~\ref{fig:resnet110_kappa},\ref{fig:resnet50_kappa} and the whiteness Figs.~~\ref{fig:resnet110_rho},\ref{fig:resnet50_rho} it is evident that the whitening methods obtain a consistently higher rank and higher whiteness than the baseline methods. Considering the accuracy figures, it is noticeable that the whitening methods converge faster, and are more stable and less noisy. The converged accuracy of the baseline, \ac{EVD} and recursive whitening methods is respectively  $73.0\%$, $73.5\%$ and $73.0\%$ for ResNet-110 and $75.7\%$, $75.9\%$ and $76.0\%$ for ResNet-50. I.e., the whitening methods are on par or slightly better than the baseline method.

\vspace{-5 pt}
\section{Conclusion}
\label{sec:conclusion}
\vspace{-5 pt}
Two novel methods for whitening the features, named \ac{EVD} and recursive gradient-based feature whitening, have been proposed. The methods offer  reduced complexity compared to the \emph{direct} feature whitening method~\cite{desjardins2015natural}, by applying the transformation to the weight gradients instead of to the activations and their gradients. The recursive method further saves computations by replacing the \ac{EVD}-based transformation with a recursive transformation, updated in steps, treating only one subspace per step and designed to gradually reduce the condition number of the features covariance-matrix. The proposed methods are applied too ResNet-110 and ResNet-50 and obtain state of the art convergence in terms of speed, stability and accuracy.

\bibliographystyle{plain}
\bibliography{refs}

\end{document}